\documentclass[conference]{IEEEtran}
\IEEEoverridecommandlockouts

\usepackage{cite}
\usepackage{amsmath,amssymb,amsfonts}
\usepackage{graphicx, enumitem}
\usepackage{textcomp}
\usepackage[ruled,vlined,linesnumbered]{algorithm2e}
\usepackage[table,dvipsnames]{xcolor}
\usepackage{tikz}
\usepackage{bm}
\usepackage{makecell}
\usepackage{booktabs}
\usepackage{multirow}
\usepackage{url}
\usepackage{tikz}
\usepackage[none]{hyphenat}
\usepackage{caption}
\usepackage{subfig}
\usepackage{comment}
\usepackage{dblfloatfix}
\usepackage{flushend}
\usepackage{dsfont} 
\usepackage{comment}

\newcolumntype{L}[1]{>{\raggedright\let\newline\\\arraybackslash\hspace{0pt}}m{#1}}
\newcolumntype{C}[1]{>{\centering\let\newline\\\arraybackslash\hspace{0pt}}m{#1}}
\newcolumntype{R}[1]{>{\raggedleft\let\newline\\\arraybackslash\hspace{0pt}}m{#1}} 

\def\BibTeX{{\rm B\kern-.05em{\sc i\kern-.025em b}\kern-.08em
    T\kern-.1667em\lower.7ex\hbox{E}\kern-.125emX}}

\begin{document}



\title{Enhanced Generalization through Prioritization and Diversity in Self-Imitation Reinforcement Learning over Procedural Environments with Sparse Rewards}

\author{
\IEEEauthorblockN{
    Alain Andres\IEEEauthorrefmark{1}\IEEEauthorrefmark{2}\IEEEauthorrefmark{3}, Daochen Zha\IEEEauthorrefmark{4}, and Javier Del Ser\IEEEauthorrefmark{2}\IEEEauthorrefmark{3}
}
\IEEEauthorblockA{\IEEEauthorrefmark{2} TECNALIA, Basque Research and Technology Alliance (BRTA), 48160 Derio, Bizkaia, Spain}

\IEEEauthorblockA{\IEEEauthorrefmark{3} University of the Basque Country (UPV/EHU), 48013 Bilbao, Bizkaia, Spain}

\IEEEauthorblockA{\IEEEauthorrefmark{4} Rice University, Houston, Texas, United States}

\IEEEauthorblockA{\IEEEauthorrefmark{1} Corresponding author: alain.andres@tecnalia.com}}

\maketitle

\IEEEpubidadjcol 
\IEEEpubid{\begin{minipage}{\textwidth}\ \\[50pt]
   \hrule height 0.5pt 
   \vspace{3pt}       
    Accepted for presentation at the IEEE Symposium on Adaptive Dynamic Programming and Reinforcement Learning (IEEE ADPRL), 2023.  \vspace{1pt}       
\end{minipage}}


\begin{abstract}
Exploration poses a fundamental challenge in Reinforcement Learning (RL) with sparse rewards, limiting an agent's ability to learn optimal decision-making due to a lack of informative feedback signals. Self-Imitation Learning (self-IL) has emerged as a promising approach for exploration, leveraging a replay buffer to store and reproduce successful behaviors. However, traditional self-IL methods, which rely on high-return transitions and assume singleton environments, face challenges in generalization, especially in procedurally-generated (PCG) environments. Therefore, new self-IL methods have been proposed to rank which experiences to persist, but they replay transitions uniformly regardless of their significance, and do not address the diversity of the stored demonstrations. In this work, we propose tailored self-IL sampling strategies by prioritizing transitions in different ways and extending prioritization techniques to PCG environments. We also address diversity loss through modifications to counteract the impact of generalization requirements and bias introduced by prioritization techniques. Our experimental analysis, conducted over three PCG sparse reward environments, including MiniGrid and ProcGen, highlights the benefits of our proposed modifications, achieving a new state-of-the-art performance in the MiniGrid-MultiRoom-N12-S10 environment. 
\end{abstract}


\section{Introduction} \label{sec:intro}
Exploration is a fundamental challenge in Reinforcement Learning (RL), especially in scenarios with sparse rewards where the agent may struggle to learn optimal decision-making due to a lack of informative feedback signals~\cite{zhang_made_2021,zhang_noveld_2022,hester_deep_2017,vecerik_leveraging_2018,nair_overcoming_2018}. 

One promising approach to improve exploration is self-Imitation Learning (self-IL), which uses a replay buffer to store past successful behaviors. In this way, the agent can reproduce and exploit rare instances of good exploration.

This idea has traditionally been achieved by selecting high-return transitions that provide positive advantage~\cite{schaul_prioritized_2016,hester_deep_2017,vecerik_leveraging_2018,nair_overcoming_2018,oh_self-imitation_2018,guo_generative_2018}. However, they often assume that agents operate in singleton environments, using the same environment for training and testing. Yet, recent studies have unveiled that this approach's susceptibility to overfitting hampers its generalization capabilities~\cite{rajeswaran_towards_2017, zhang_dissection_2018}. To address this issue, the adoption of procedurally-generated (PCG) environments has been proposed~\cite{chevalier-boisvert_maxime_minimalistic_2018}, where a different environment is generated in each episode. Unfortunately, self-IL approaches often yield poor results in PCG environments, as the agent may not be able to encounter a high-return transition more than once~\cite{zha_rank_2021}.

In previous work we proposed \textbf{Ra}nking the E\textbf{pi}so\textbf{d}es (RAPID) \cite{zha_rank_2021}, which takes into account the whole episode rather than isolated transitions, effectively distinguishing good exploration behaviors that handle state space changes affecting the pursued generalization capacity. We discovered that episode-level selection can significantly boost the sample efficiency in PCG environments~\cite{zha_rank_2021}.
However, RAPID solely focuses on which experiences to store, and replays data uniformly from the buffer, indirectly implying that all experiences of such self-collected demonstrations are equally valuable. Furthermore, it does not guarantee diversity of behaviors between the stored episodes~\cite{andres_towards_2022,raileanu_decoupling_2021}, potentially resulting in over-fitted policies incapable of generalizing.

To address the above limitations, in this work, we propose methods to specifically tailor self-IL sampling strategies (Section \ref{subsec:buffer_sampling}). Instead of treating all transitions as equally significant, we prioritize them in different ways. Moreover, we extend prioritization techniques that have been previously evaluated with off-policy methods in singleton environments to PCG environments. 
Furthermore, we propose modifications to counteract the diversity loss stemming from the tasks' generalization requirements and the bias introduced by prioritization techniques (Section \ref{subsec:buffer_persistence}). The resulting analysis demonstrates the performance benefits of our modifications through experiments conducted in three PCG sparse reward environments: \texttt{MultiRoom} and \texttt{ObstructedMaze} from MiniGrid ~\cite{chevalier-boisvert_maxime_minimalistic_2018} and \texttt{Ninja} from ProcGen ~\cite{cobbe_leveraging_2020} (Section \ref{sec:exp}). In particular, our approach establishes a new state-of-the-art in the \texttt{MiniGrid-MultiRoom-N12-S10} environment.

\section{Preliminaries} \label{sec:preliminaries}

This section begins with a background on self-Imitation Learning (Section~\ref{sec:exploration}), followed by a discussion regarding the Experience Replay Buffer adoption (Section~\ref{sec:exploration}). Then we formally describe our research goal (Section~\ref{subsec:researchobjective}).

\subsection{Self-Imitation Learning} \label{sec:exploration}

Self-imitation learning (self-IL) is a promising approach to foster exploration by leveraging good behaviors exhibited in the past. Initially introduced in the so-called SIL~\cite{oh_self-imitation_2018}, this method employs a replay buffer to retain historical state-action pairs, imitating only those pairs that yielded greater returns than the agent's value estimate in previous episodes. By leveraging good past behaviors, the agent achieves enhanced exploration capabilities, ultimately leading to improved sample efficiency~\cite{oh_self-imitation_2018}.

However, SIL ~\cite{oh_self-imitation_2018} assumes that agents operate solely within singleton environments, meaning that the same environment is employed for both training and testing purposes. Recent studies have revealed that training an agent in such a manner renders it susceptible to overfitting, hindering its ability to generalize~\cite{rajeswaran_towards_2017, zhang_dissection_2018}. Thus, the utilization of PCG environments has been advocated~\cite{chevalier-boisvert_maxime_minimalistic_2018}. By generating distinct environments for each episode, the agent is encouraged to learn generalizable skills.

To enable self-IL in PCG environments, we proposed RAPID in our previous work~\cite{zha_rank_2021}. The idea behind RAPID is to facilitate agents to replicate good episode-level exploration behaviors. This is accomplished by assigning a score to the entire trajectory and rank all the past trajectories in a small replay buffer, based on the following formulation:
\begin{equation}\label{eq:rapid_scores}
    S = w_0 \cdot S_{ext} + w_1 \cdot S_{local} + w_2 \cdot S_{global},
\end{equation}
where $S_{ext}$ determines the extrinsic Monte Carlo return, $S_{local}$ promotes diversity of states within the episode, $S_{global}$ fosters the lifelong training exploration, and all the $w_*$ are used to balance the weight given to each score. Building upon RAPID, our previous work further introduced Intrinsic Motivation (IM) to enable better exploration~\cite{andres_towards_2022}.

While RAPID~\cite{zha_rank_2021} and RAPID+IM~\cite{andres_towards_2022} have achieved promising performance, their utilization of a strict ranking-based buffer and uniform replay strategy may result in certain state-action pairs overpowering the learning process. This can potentially lead to learning divergence, as exemplified in the \texttt{MultiRoom-N12-S10} environment~\cite{zha_rank_2021}.

\subsection{Experience Replay Buffer} 

One pivotal component of self-IL is the Experience Replay Buffer~\cite{lin_self-improving_1992, mnih_human-level_2015}, a data structure designed to retain past experiences. This buffer empowers the agent to effectively reuse previous experiences to enhance learning efficiency. In the past, numerous strategies have been proposed to enhance experience replay. These strategies can be broadly categorized into two groups: 1) defining how to replay experiences from the buffer, and 2) determining which experiences to store.

\subsubsection{How to replay} \label{subsec:rl_replay}

Various non-uniform reply strategies have emerged, such as using the TD-error as proxy \cite{schaul_prioritized_2016}, applying importance sampling~\cite{schaul_prioritized_2016,de_bruin_experience_2018}, minimizing meaningless updates with episode-level sampling~\cite{lee_sample-efficient_2019}, adopting hierarchical experience replay~\cite{yin_knowledge_2017}, sampling frequently visited transitions~\cite{sun_attentive_2020}, and optimizing the use of the buffer with meta-learning approaches~\cite{zha_experience_2019, oh_learning_2021}.

\subsubsection{Which experiences to store} \label{subsec:rl_store}

The most fundamental approach is the First-In-First-Out (FIFO) strategy, which employs a fixed-size memory to sequentially store data~\cite{mnih_human-level_2015}. The works in ~\cite{zhang_deeper_2018,de_bruin_experience_2018,fedus_revisiting_2020} uncovered the significant impact of different sizes of Experience Replay Buffers on performance. Subsequently, a range of strategies has been proposed to enhance its utility. These strategies include increasing the diversity of experiences employing short-term and long-term buffers~\cite{de_bruin_importance_2015, de_bruin_improved_2016,isele_selective_2018}, prioritizing the storage of experiences with high rewards in a greedy manner~\cite{karimpanal_experience_2018}, continuously refreshing the buffer according to the current policy~\cite{du_lucid_2021}, utilizing multiple buffers representing different event types ~\cite{kompella_event_2023,wurman_outracing_2022}, as well as employing Hindsight Experience Replay~\cite{hong_topological_2022}.

\subsection{Research Objective} \label{subsec:researchobjective}

Despite the aforementioned efforts, none of the strategies to manage the Experience Replay Buffer were specifically designed for self-IL, nor have they been tested in PCG environments. There lies the research goal pursued in this work: to examine the impact of various experience replay strategies for self-IL approaches within PCG environments. Our objectives to accomplish this goal are threefold: 1) to assess the efficacy of prioritization strategies in replay mechanisms, 2) to empirically evaluate the effectiveness of filtering strategies to avoid meaningless updates, and 3) to explore whether improving data diversity in the buffer could contribute positively to the learning process.

\section{Designed strategies for Efficient Experience Replay in Self-Imitation Learning}

We now introduce methods to prioritize the sampling from the buffer (Section \ref{subsec:buffer_sampling}) and modifications to promote the diversity of transitions therein stored (Section \ref{subsec:buffer_persistence}).

\subsection{Prioritization \& Filtering} \label{subsec:buffer_sampling}

Uniform sampling strategy has been largely adopted in experience replay due to its simplicity. However, an agent could learn more effectively from some transitions than from others. Motivated by this, we designed several prioritization and filtering methods for self-IL.

\subsubsection{Prioritization}
The idea is to replay some experiences with more frequency due to their significance in learning. We extend the idea of PER ~\cite{schaul_prioritized_2016} considering three different proxies for \textit{prioritization}: 
\begin{itemize}[leftmargin=*]

    \item \textbf{TD-error} $\rightarrow \delta=r_t + \gamma \cdot V(s_{t+1}) - V(s_t)$, where $r_t$ is the reward, $V(\cdot)$ is the value function, $\gamma$ is the discount factor. It represents how \textit{surprising} or unexpected a given transition is with respect to the knowledge retained at the agent. It has been shown to work well in practice with algorithms that have already computed the TD-error for updating its parameters (e.g., Q-learning or SARSA ~\cite{watkins_q-learning_1992,rummery_-line_1994,mnih_human-level_2015}). However, it can be a poor estimate under some circumstances, such as partial observability, sparse rewards, and stochastic transitions \cite{schaul_prioritized_2016}, which are common in PCG environments.
    
    \item \textbf{Log-Likelihood} $\rightarrow \sum_{(s_t, a_t) \sim B} ln(\pi(a_t|s_t))$, where $\pi(a_t|s_t)$ is the probability of selecting action $a_t$ given the current state $s_t$. It suggests how likely an action may occur in a given state. The log-likelihood prioritization promotes frequent actions.
    
    \item \textbf{Novelty} $\rightarrow 1/\sqrt{N(s_{t})}$, where $N(s_{t})$ stands for the state visitation counts throughout the whole training. It aims to promote transitions that are more novel, fostering the exploration in those states in which the agent is uncertain about its captured knowledge.   
\end{itemize}

Once the score $p_i$ of each transition has been calculated according to any of the above proxies, the probability $P(i)$ of sampling a given transition $i$ is defined as:
\begin{equation}\label{eq:sampling_probab}
    P(i) = \frac{p_i^\alpha}{\sum_k p_k^\alpha},
\end{equation}
where $\alpha$ is a hyperparameter used to determine how much prioritization is applied\footnote{In Expression (2), $\alpha=0$ refers to the uniform sampling case, whereas $\alpha=1$ stands for maximum prioritization.}.

\subsubsection{Filtering}

Instead of prioritizing the experiences based on different proxies/scores like those explained above, an alternative strategy is to sample uniformly from the buffer, but apply some \textit{filters} to avoid undesired updates. We consider the following filtering objectives:
\begin{itemize}[leftmargin=*]
    \item \textbf{Non-zero Return Trajectories} $\rightarrow G_t > 0$, where the discounted return is given by $G_t = \sum_{k=0}^\infty \gamma^k r_{t+k}$. It grants priority to those trajectories that represent a valid/success example\footnote{We refer as valid success examples to any trajectory with $G_t\neq 0$.} to complete the task. When no success trajectories are available in the buffer, the agent samples them uniformly. However, when a valid demonstration exists, the agent will imitate those experiences greedily.
    
    \item \textbf{Positive Advantage} $\rightarrow \left[G_t - V(s_t)\right]_{+}$, where $G_t$ is the discounted return, $V(\cdot)$ is the value function, and $[\cdot]_+ = \max(\cdot,0)$. Akin to SIL \cite{oh_self-imitation_2018}, this option only considers experiences that are expected to have a positive impact on the agent's learning process. Therefore, if a $\{s,a\}$ tuple has a worse return ($G_t$) than the one expected by the agent ($V(s_t)$), that transition is not considered for replay. 

    \item \textbf{Unique states} $\rightarrow \mathds{I}\{N_e(s_{t+1})\}$, where $\mathds{I}\{\cdot\}$ takes value $1$ if its argument is $1$ ($0$ otherwise), and $N_e(s_{t+1})$ stands for the state visitation counts within an episode. Motivated by episodic exploration success approaches \cite{andres_evaluation_2022,henaff_study_2023,wang_revisiting_2023}, with this filter w foster exploration by preventing the replay of transitions in the same trajectory that lead to the same $s_{t+1}$.

\end{itemize}

\subsection{Data Diversity} \label{subsec:buffer_persistence}
RAPID focused exclusively on episode-level scores, without considering whether the stored demonstrations are closely related and provide a comprehensive representation of the required diversity. In previous studies conducted in PCG environments, it was observed that optimal solutions -- and $V(s_t)$ -- can significantly vary from one level to another, even within the same task \cite{andres_towards_2022,raileanu_decoupling_2021}. As a result, certain levels may yield a superior extrinsic return, $G_t$, which can inadvertently bias the diversity of the trajectories considered by RAPID, and lead to behavioral overfitting. Furthermore, the data replay strategies introduced earlier can induce an additional bias toward the selected proxy. We propose two strategies to alleviate such issues:
\begin{itemize}[leftmargin=*]
\item \textbf{On-policy Novelty} (\textit{Intrinsic Motivation}): The on-policy updates are decoupled from the off-policy ones. Therefore, novelty-seeking techniques such as Intrinsic Motivation \cite{zhang_made_2021,zhang_noveld_2022} can be used not only to promote the exploration, but also as a tool to prevent the agent from getting stuck in a local optimum due to low-diversity off-policy updates.
    
\item \textbf{Forced Diversity}. Instead of considering any episode belonging to any level, we constrain the buffer so that it always contains a fixed number of episodes per level, \textit{forcing} diversity among the levels represented in the buffer.
\end{itemize}

\section{Experiments \& Results} \label{sec:exp}

This section describes the experimental setup used to assess the performance of the aforementioned strategies. Specifically, the environments and selected hyperparameter values are given in Section \ref{subsec:exp_setup}, whereas results are analyzed and discussed in Section \ref{subsec:results}. 

\subsection{Experimental Setup}\label{subsec:exp_setup}
\paragraph{Environments} \label{subsec:exp_env}
Performance evaluations are carried out over MiniGrid and Procgen PCG environments, where the agent position, background, and even configuration of objects randomly change from episode to episode. We only consider sparse reward tasks that constitute an exploration challenge:
\begin{itemize}[leftmargin=*]
\item \textbf{MiniGrid} \cite{chevalier-boisvert_maxime_minimalistic_2018}: 
Within this benchmark, we evaluate the solutions over a MultiRoom environment with 12 rooms and a maximum room size of 10 --\texttt{MN12S10}-- where the agent has to open doors and move forward until reaching a far green square goal. In addition, we also consider an ObstructedMaze scenario -- \texttt{O1Dlhb} -- in which the agent must find out where a hidden key is, move a ball that obstructs the opening of the locked door, and move forward to another room where the goal is accomplished when picking up the ball placed in it. The agent is fed with a partially observable state of dimensions $7 \times 7 \times 3$ that represents the surroundings in a compact manner. The agent is capable of executing up to 7 possible discrete actions.

\item \textbf{ProcGen} \cite{cobbe_leveraging_2020}:
Among the 16 possible environments, we opt for \texttt{Ninja} as the agent is only rewarded with either a +0 or +10 reward depending on whether the agent succeeds in the completion of the task. The agent has to move forward while avoiding bombs that can kill itself, so it has to jump and move through multiple elevated platforms carefully. The input observation consists of a $64 \times 64 \times 3$ image, while the action space consists of 15 possible discrete values.
\end{itemize}

\paragraph{Hyperparameters} \label{subsec:baselines_hyperparams}
For MiniGrid, we use the hyperparameters and neural network architectures proposed in \cite{andres_towards_2022}. That is, we use Proximal Policy Optimization (PPO) \cite{schulman_proximal_2017} with an actor-critic framework, using 64-64 Multi-Layer Perceptron. Besides, we select the best value resulting from a grid search for $\alpha \in [0.2, 0.4, 0.6, 0.8, 1.0]$ in Equation \eqref{eq:sampling_probab} when using any of the prioritizing strategies explained in Section \ref{subsec:buffer_sampling}. That is, we use$\alpha=0.6$ in \texttt{MN12S10} and $\alpha=1.0$ in \texttt{O1Dlhb} for \textit{Novelty} prioritization; for the rest of cases, we opt for $\alpha=0.2$. As for ProcGen, we adopt the hyperparameters and architectures used in \cite{cobbe_leveraging_2020}. Furthermore, when IM is employed, we use BeBold \cite{zhang_noveld_2022} as in \cite{andres_towards_2022}.

\subsection{Results and Discussion} \label{subsec:results}

We report the mean and standard deviation of the average return calculated over the last 100 episodes for each experiment. Such statistics are computed over 3 different runs to account for the statistical variability of the results. In addition, we consider two batch sizes ($\mathcal{B}_{IL}$) for self-IL; unless otherwise stated, solid curves represent results obtained with $\mathcal{B}_{IL}=256$, whereas dash-dotted lines indicate that the batch size in use is $\mathcal{B}_{IL}=2048$. 
\begin{figure}[!h]
    \centering
    \includegraphics[width=0.45\textwidth]{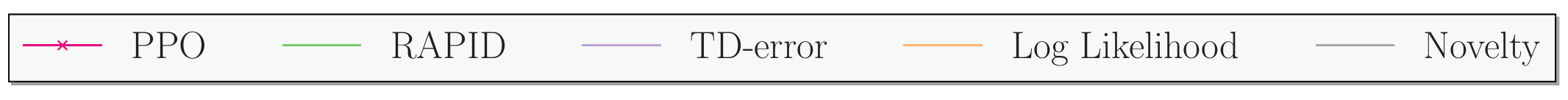} 
    \includegraphics[width=0.45\textwidth]{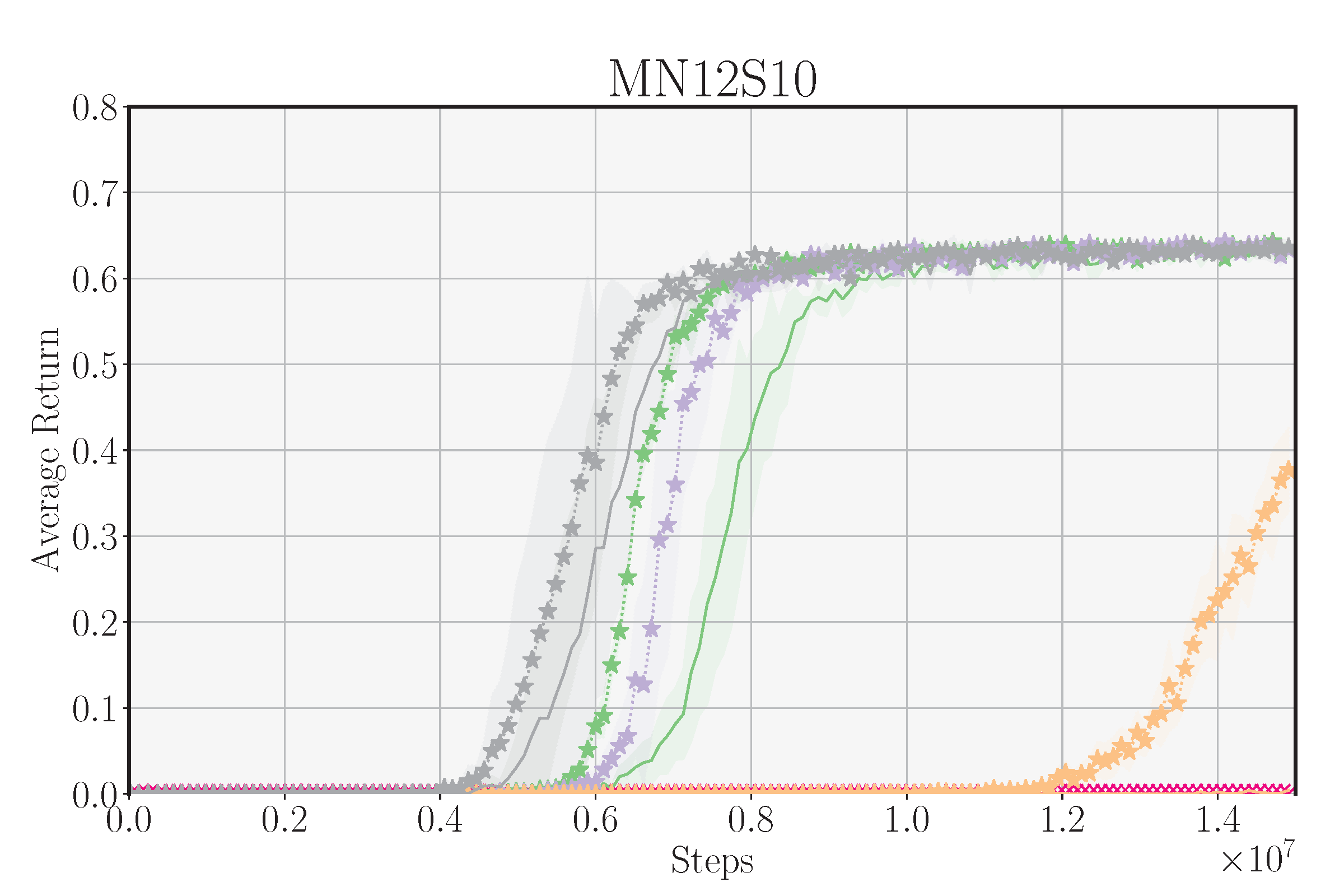}
    \includegraphics[width=0.45\textwidth]{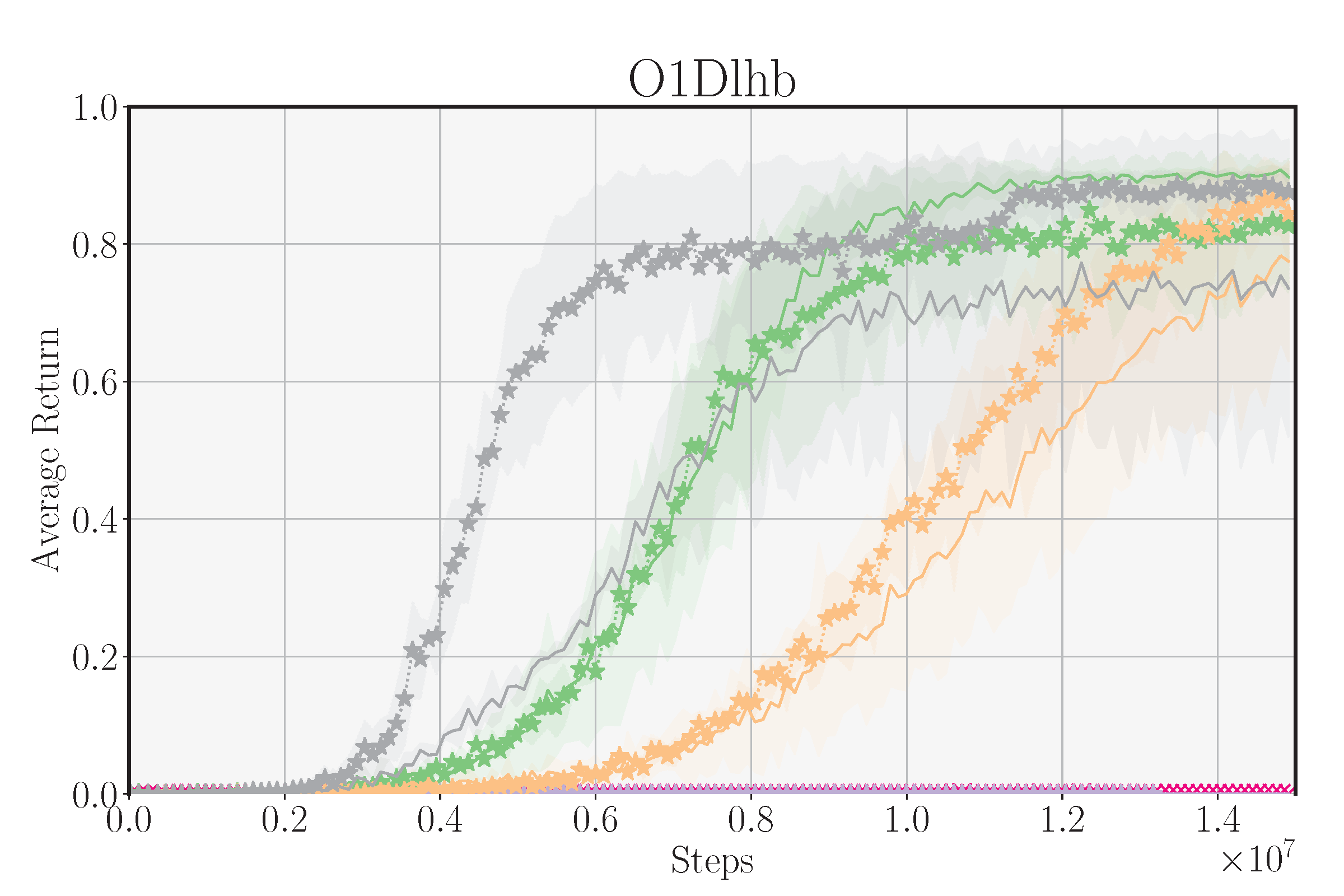}
    \\
    \caption{Performance of the agent when adopting prioritization strategies in \texttt{MN12S10} and \texttt{O1Dlhb} tasks.}
    \label{fig:prioritization}
\end{figure}

\subsubsection{Prioritization \& Filtering} \label{subsubsec:results_replay}

\begin{figure}[!h]
    \centering    \includegraphics[width=0.45\textwidth]{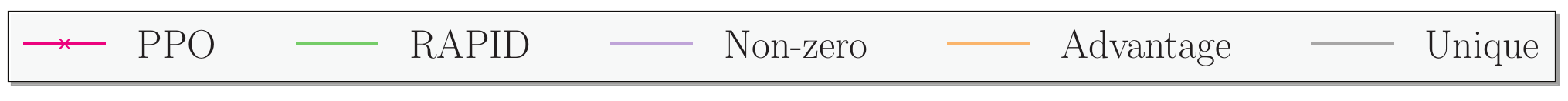}
    \includegraphics[width=0.45\textwidth]{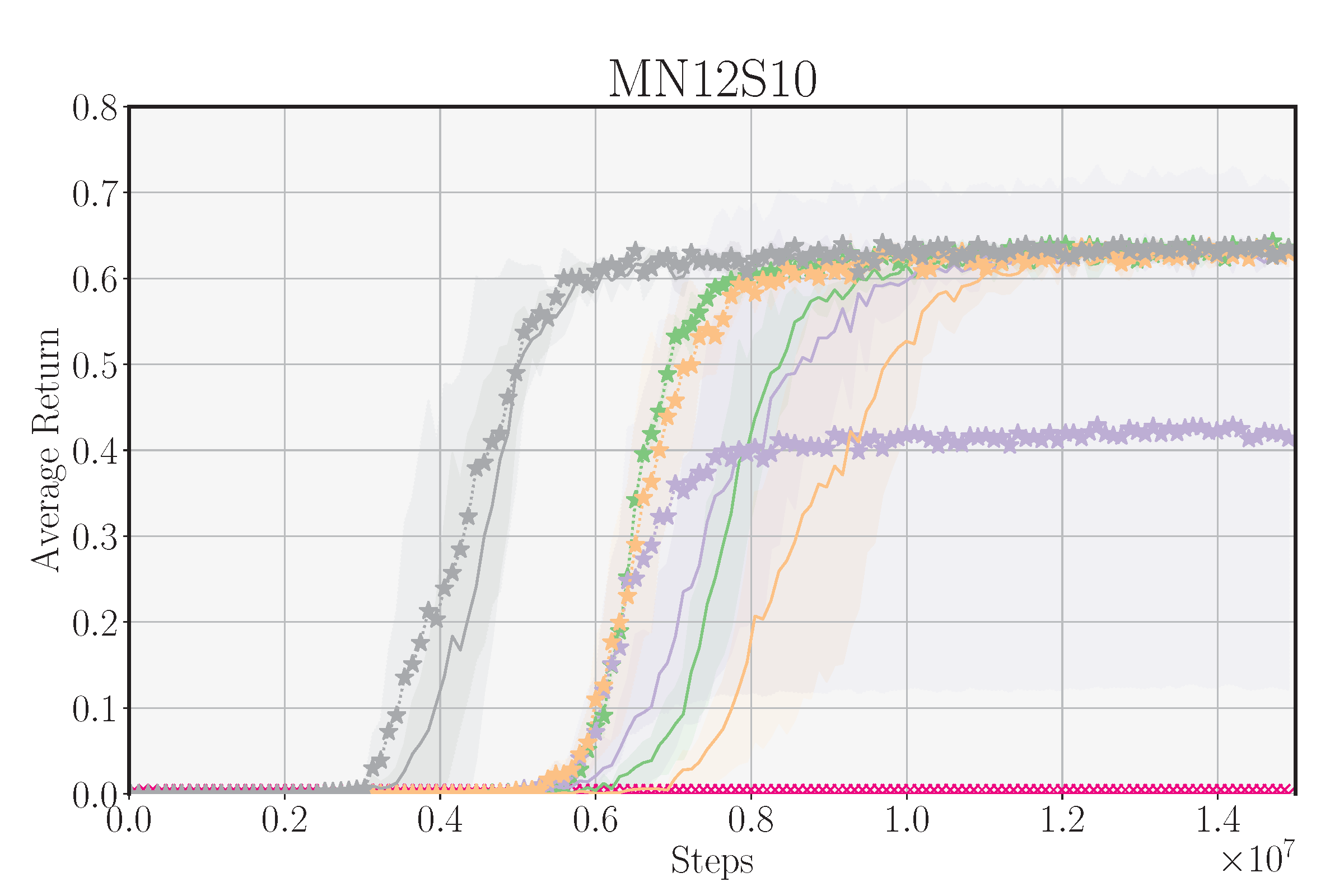}
    \includegraphics[width=0.45\textwidth]{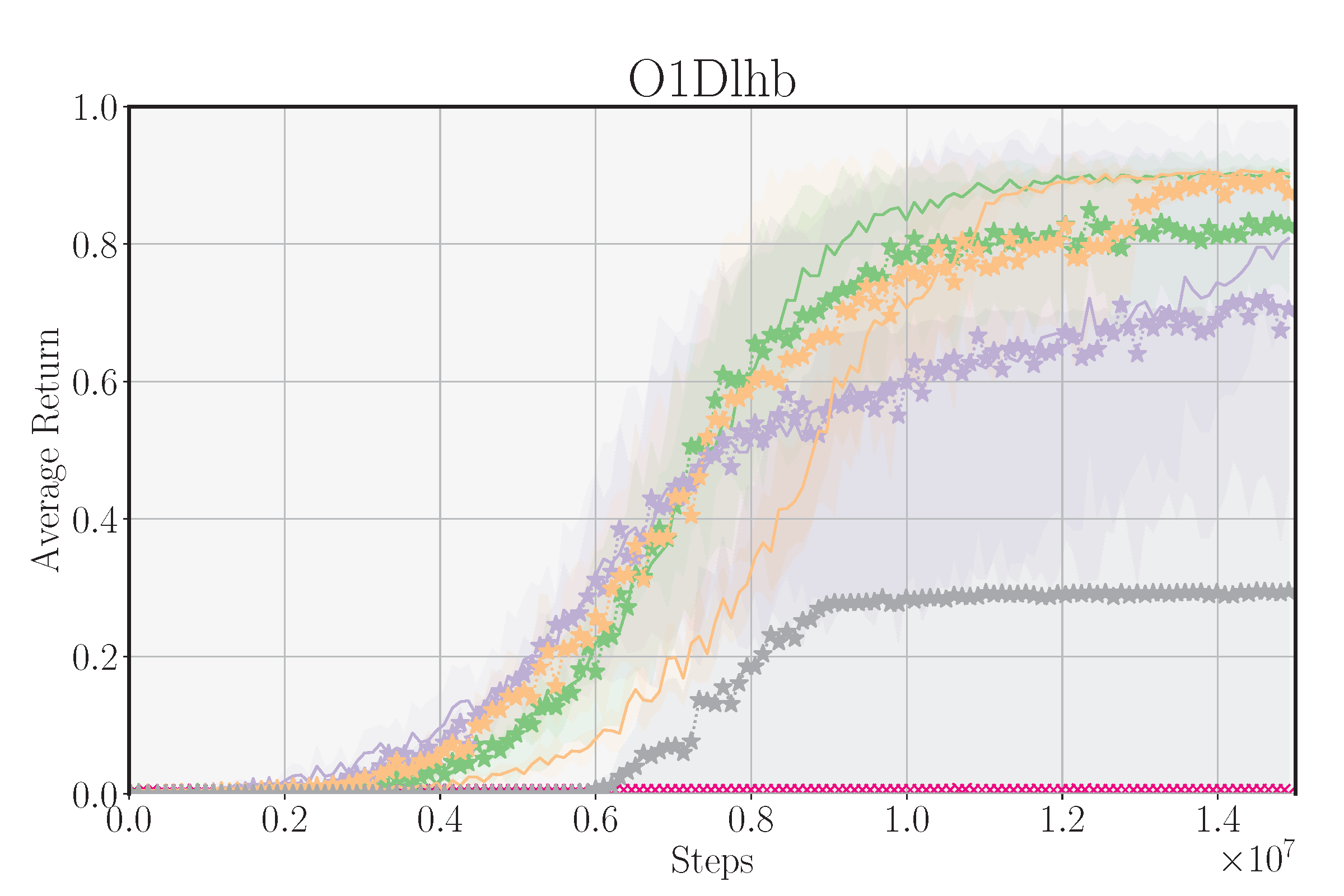}
    \caption{Performance of the agent when adopting filtering strategies in \texttt{MN12S10} and \texttt{O1Dlhb} tasks.}
    \label{fig:filtering}
\end{figure}

According to the results in Figures \ref{fig:prioritization} and \ref{fig:filtering}, the following observations can be made:

Firstly, \textbf{prioritization} (Figure \ref{fig:prioritization}) based on \textit{Novelty} (gray) consistently outperforms uniform sampling (green) in terms of sample efficiency for learning optimal policies. By contrast, using \textit{TD-error} (magenta) as a proxy for prioritization yields poor results, only enabling learning over \texttt{MN12S10} with large batch sizes and failing in other cases. Moreover, the proposed \textit{Log-Likelihood} prioritization strategy (yellow) manages to learn a good policy, but requires more interactions than uniform sampling.

When it comes to \textbf{filtering} strategies, Figure \ref{fig:filtering} reveals that these methods render faster learning compared to uniform sampling (green). In fact, the best result on \texttt{MN12S10} is achieved with the \textit{Unique} filtering strategy. However, it is worth noting that adopting these filtering methods can potentially lead to a loss of diversity, as the agent tends to overfit to a subset of the entire experience distribution (e.g., \textit{Non-zero}, magenta color, gets stuck in local optima solutions despite beginning to learn earlier on training). In contrast, uniform sampling (green) maintains diversity but at the cost of a slower learning and a lower sample efficiency. 

Last but not least, increasing the \textbf{batch size} ($\mathcal{B}_{IL}=2048$) improves sample efficiency and the likelihood of obtaining a successful policy compared to a smaller batch size ($\mathcal{B}_{IL}=256$). This trend is consistent across all scenarios and cases, where the agent converges faster and/or achieves a valid policy. Increasing the batch size ensures that larger amounts of information are considered within each update, maximizing the probability of receiving valuable feedback and minimizing variance. This aspect is particularly critical in PCG environments, where generalization is essential for effective learning and exploration \cite{andres_towards_2022}. However, it is important to note that while increasing the batch size reduces learning variance, it may also introduce a trade-off by potentially increasing bias. This trade-off is observed in scenarios such as \texttt{MN12S10} with the \textit{Non-zero} filtering strategy and in \texttt{O1Dlhb} across all algorithmic approaches, where a larger batch size leads to faster learning, but poses challenges in attaining the optimal expected performance.

Based on these observations, it is hypothesized that by ensuring diversity while employing the aforementioned prioritization and filtering techniques, both sample efficiency and the agent's learning capacity can be improved.

\vspace{3mm}
\subsubsection{Analysis of Data Diversity} \label{subsubsec:results_persistence}

\begin{figure}[!h]
    \centering
    \includegraphics[width=0.9\columnwidth]{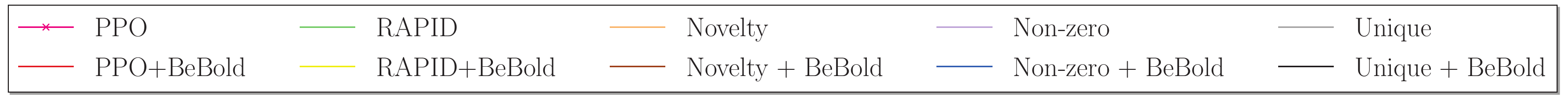}\\
    \includegraphics[width=0.9\columnwidth]{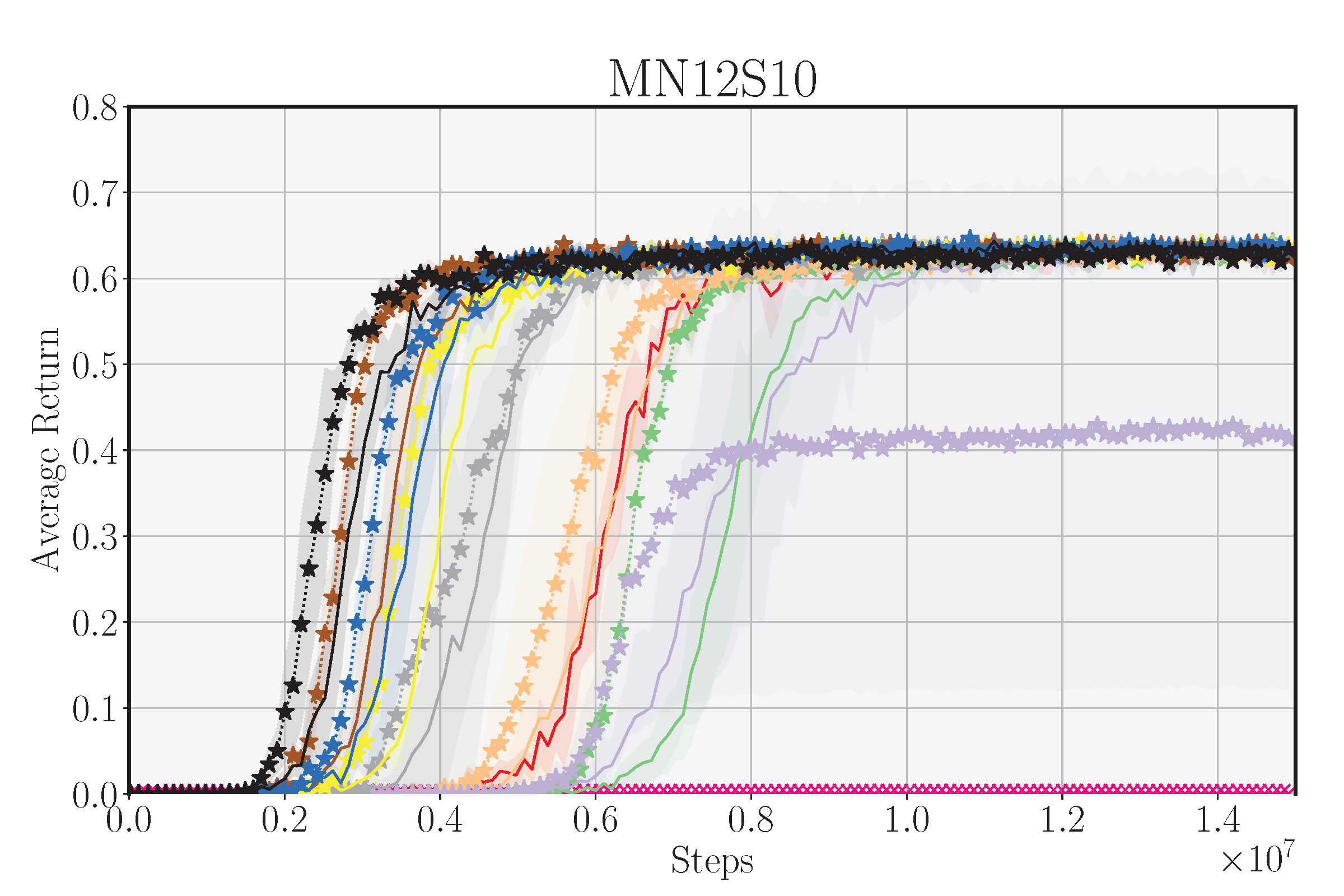}\\
    \includegraphics[width=0.9\columnwidth]{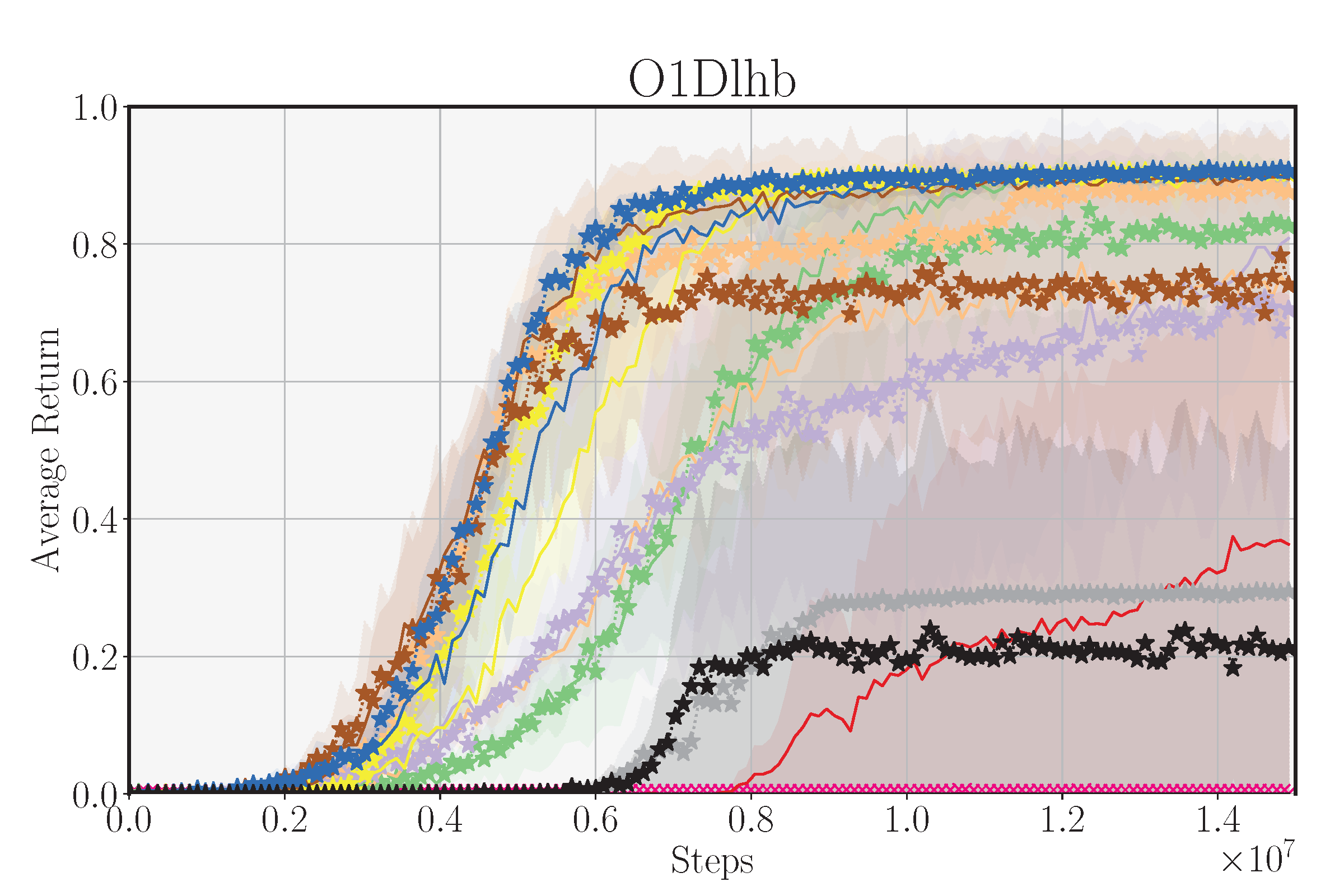}
    \caption{
    Performance of the agent when fostering the diversity of self-IL with IM while adopting prioritization and filtering strategies in \texttt{MN12S10} and \texttt{O1Dlhb} tasks. 
    }
    \label{fig:bebold_sil}
\end{figure}

\vspace{1mm}\paragraph{Intrinsic Motivation} In prior work \cite{andres_towards_2022}, the combination of self-IL and IM in PCG environments was demonstrated to be a successful framework for improving sample efficiency. However, the influence of IM on diversity remained unexplored. Figure \ref{fig:bebold_sil} exposes that \textbf{prioritization and filtering methods that previously resulted in suboptimal performance} (e.g., \textit{non-zero}) \textbf{can now converge to the optimal policy when utilizing BeBold}.

More importantly, the proposed modifications poses a new landmark performance level in terms of sample efficiency \cite{andres_towards_2022}. This is evident when comparing the number of samples required to reach a specific performance level. For instance, in \texttt{MN12S10}, the \textit{Novelty} prioritization (brown) and \textit{Unique} filtering (black) methods converge to the expected optimal policy in approximately $3.8M$ steps, while uniform sampling (yellow) requires around $5M$ steps\footnote{Measures taken for curves with $\mathcal{B}_{IL}=2048$. Improvements are also consistent for $\mathcal{B}_{IL}=256$.}. That is, \textbf{the same optimal solution is achieved with $\mathbf{24}$\% fewer interactions}. Similarly, the \textit{Non-zero} filtering (blue) approach improves sample efficiency by approximately $10\%$. In the case of \texttt{O1Dlhb}, sample efficiency gaps are less noticeable. However, the \textit{Non-zero} filtering approach (blue) still dominates by \textbf{reducing the number of required steps by about $\mathbf{11}$\%}.

\begin{figure}[h]
    \centering
    \includegraphics[width=\columnwidth]{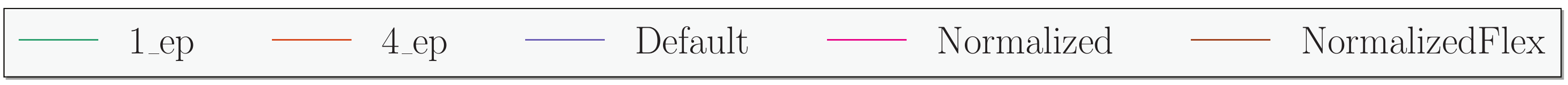}
    \includegraphics[width=\columnwidth]{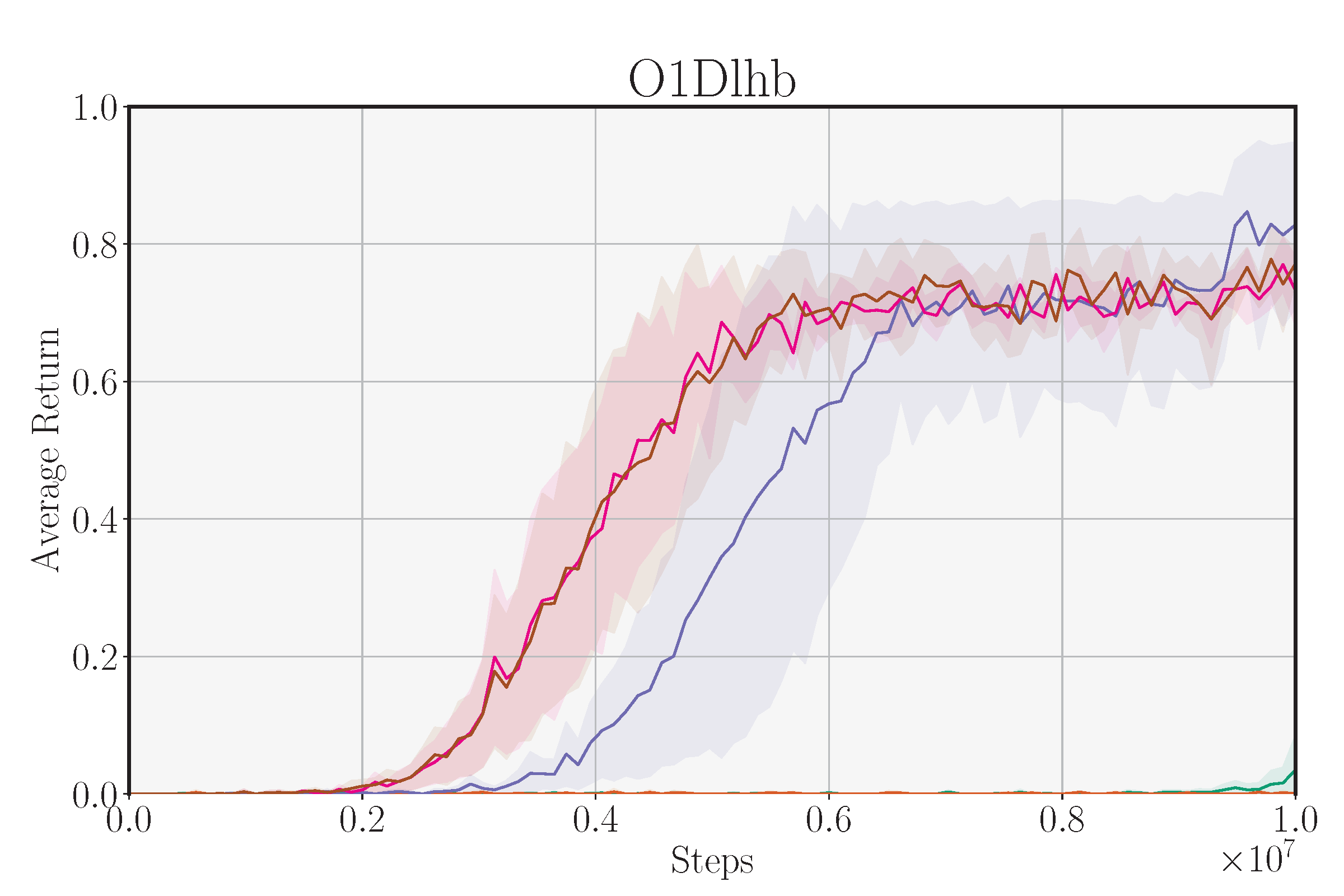}
    \caption{Average train return over 10 levels in \texttt{O1Dlhb}. The expected optimal performance is approximately $0.9$. All the shown curves use RAPID+IM (i.e., BeBold) and $\mathcal{B}_{IL}=256$.}
    \label{fig:o1dlhb_10levels_diversity}
\end{figure}

\paragraph{Forced Diversity} We recall that in Section \ref{subsec:buffer_persistence} we indicate the potential need for addressing the lack of diversity in the stored trajectories. In light of the results reported in Section \ref{subsubsec:results_replay}, the diversity should be addressed. One way to do this is by imposing a fixed number of episodes per level (i.e., 1 episode per level, $1\_ep$, or 4 episodes per 
 level, $4\_ep$), so that the representation of all levels in the buffer is always guaranteed. We evaluate this idea in Figure \ref{fig:o1dlhb_10levels_diversity}, with 10 training levels for task \texttt{O1Dlhb}. Neither $1\_ep$ nor $4\_ep$ achieve better results with respect to the \textit{Default} setup executed with RAPID regardless of diversity issues. 

Furthermore, we also consider two more ablation studies related to the difference of optimal solutions between levels \cite{andres_towards_2022,raileanu_decoupling_2021}: \textit{Normalized} and \textit{NormalizedFlex}. The first normalizes the extrinsic return obtained from the environment according to the optimal number of steps of each level, preventing the agent from repleting the buffer content with levels requiring shorter optimal paths. The second approach --\textit{NormalizedFlex}-- is an extension of \textit{Normalized} that assigns a maximum return score of 1.0 to any collected trajectory that requires 0 to 20 steps more than the actual shortest solution. Based on these modifications, the agent manages to learn with fewer interactions. However, \textbf{all the analyzed approaches are incapable of performing well in all the 10 selected training levels}. Concretely, they fail into learning $1$ out of $10$ levels.

\begin{figure}[t]
    \includegraphics[width=\columnwidth]{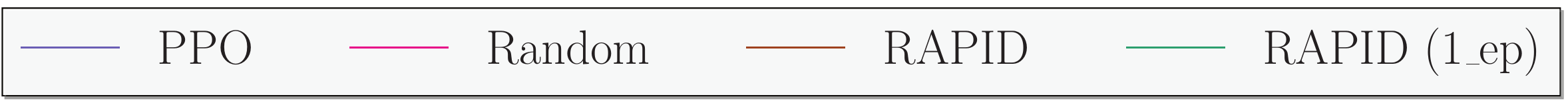}
    \includegraphics[width=\columnwidth]{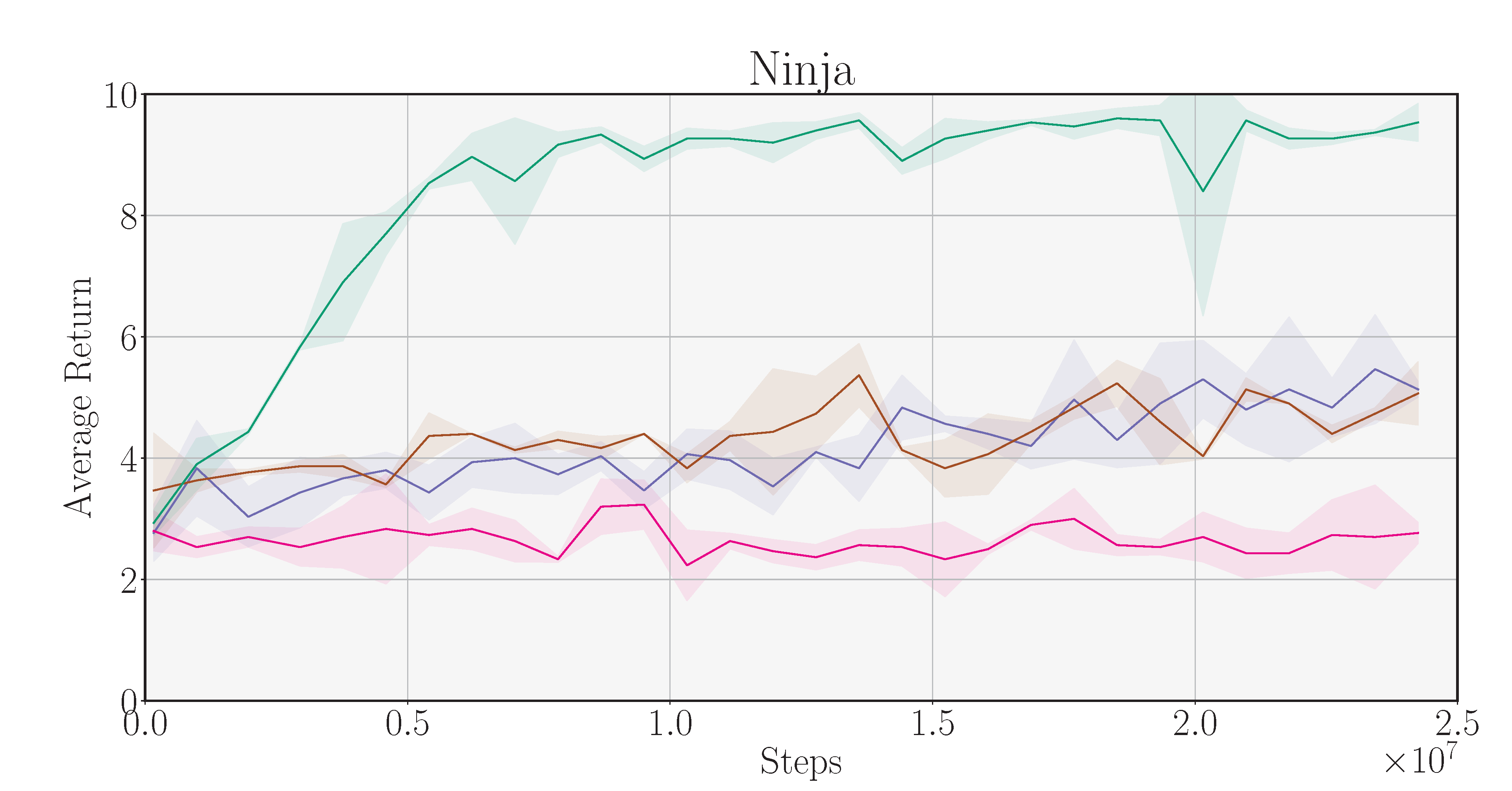}
    \caption{Average train return over 200 levels in ProcGen's \texttt{Ninja} task. RAPID is used with $w_1=w_2=0$. The expected optimal performance is equal to $10$.}
    \label{fig:ninja_200levels}
\end{figure}

For completeness, we extend the aforementioned results to another environment with different state space and complexity (\texttt{Ninja}). Such results are summarized in Figure \ref{fig:ninja_200levels}. Indeed, since the input is an image, RAPID cannot compute the local and global exploration scores, as it was originally designed for discrete state spaces \cite{zha_rank_2021}. Therefore, we applied RAPID considering only the extrinsic return, i.e., $S = w_0 \cdot S_{ext} = \sum_k \gamma^k r_{t+k}$ with respect to the original Equation \eqref{eq:rapid_scores}. As opposed to previous outcomes, we observe that \textbf{in \texttt{Ninja}, storing at least one episode per level has a positive impact} on the agent's learning, making the difference between learning an almost optimal policy or a policy that barely surpasses a random agent's performance.

\section{Conclusions and Future Work}\label{sec:conc}

The pivotal idea we wish readers to take away from this research is the threefold necessity of applying self-IL, particularly for generalization over PCG environments. 

Firstly, it is crucial to manage which demonstrations to store in the buffer, showing that episode-level exploration scores are suitable for this purpose \cite{zha_rank_2021}. Secondly, uniform sampling replay strategies assume all the stored experiences to be equally valuable, despite not being necessarily true. The prioritizing and filtering strategies proposed in this work have shown improvements in terms of sample efficiency, with \textit{Novelty} prioritization and \textit{Unique} filtering the ones yielding better outcomes. Notably, the latter has established a new state-of-the-art performance in \texttt{MN12S10}. Moreover, we have shown that increasing the imitation batch size increases the probability of sampling valuable experiences, reducing the variance of the updates and the required interactions to learn.

Last but not least, ensuring diversity is necessary for generalization. We have concluded that Intrinsic Motivation is an effective tool not only to foster on-policy exploration, but also to avoid the overfitting derived from prioritization and filtering techniques. Furthermore, in some scenarios like MiniGrid, having a diverse set of episodes to be imitated is not of help. However, in other tasks like \texttt{Ninja} from ProcGen, forcing such diversity between episodes can boost performance.

In the future, this research work can be extended by using alternative techniques that guarantee the diversity between the stored episodes. Besides Behavior Cloning, it would be interesting to see how other techniques related to Imitation Learning (e.g., GAIL, GASIL~\cite{ho_generative_2016,guo_generative_2018}) or Inverse Reinforcement Learning can perform in these PCG environments. Finally, finding out effective exploration scores suitable for continuous state spaces would allow the adoption of RAPID to learn over wider problems.   

\section*{Acknowledgments}
A. Andres and J. Del Ser receive support from the FaRADAI project (ref. 101103386) funded by the European Commission under the European Defence Fund (EDF-2021-DIGIT-R). J. Del Ser also acknowledges funding from the Basque Government (MATHMODE, IT1456-22).

\bibliographystyle{IEEEtran}
\bibliography{bibtex}

\end{document}